%% file: full_paper.tex
\documentclass{article}

\usepackage{iclr2027_conference,times}
\input{math_commands.tex}

\usepackage[utf8]{inputenc}
\usepackage[T1]{fontenc}
\usepackage{amsmath,amssymb,amsthm,mathtools}
\usepackage{bm}
\usepackage{booktabs}
\usepackage{multirow}
\usepackage{array}
\usepackage{tabularx}
\usepackage{makecell}
\usepackage{algorithm}
\usepackage{algorithmic}
\usepackage{enumitem}
\usepackage{xcolor}
\usepackage{microtype}
\usepackage{graphicx}
\usepackage{hyperref}
\usepackage{url}
\usepackage{caption}
\newtheorem{theorem}{Theorem}[section]
\newtheorem{lemma}[theorem]{Lemma}

\newtheorem{corollary}[theorem]{Corollary}
\theoremstyle{definition}
\newtheorem{definition}[theorem]{Definition}
\newtheorem{assumption}[theorem]{Assumption}
\theoremstyle{remark}

\newcommand{\Prob}{\mathbb{P}}

\newcommand{\ind}{\mathbb{I}}
\newcommand{\safe}{\mathrm{safe}}

\title{StreamLTT: Risk Certification for Policy Selection under Stream Dependence}

\author{
Jinwen Xu \\
\textit{School of ECE} \\
\textit{University of Georgia} \\
Athens, GA 30602, USA \\
\texttt{jinwen.xu@uga.edu}
\And
Qin Lu \\
\textit{School of ECE} \\
\textit{University of Georgia} \\
Athens, GA 30602, USA \\
\texttt{qin.lu@uga.edu}
}

\iclrfinalcopy

\begin{document}
\maketitle

\lhead{Preprint}

\begin{abstract}
Learn-then-Test (LTT) provides a general framework for certifying deployment
policies against a prescribed risk target, but its standard formulation
targets a fixed population risk estimated from calibration losses.
In sequential deployment, losses may instead be dependent and nonstationary,
so the risk of a fixed policy can evolve with the observed history. We
introduce \emph{StreamLTT}, which replaces this fixed population risk with
\emph{average predictable risk}, defined by conditioning each decision’s
loss on the information available before its outcome is revealed.
Conditional centering turns empirical deviations from this target into a
martingale sum, enabling both a test that adapts to predictable variance and
a finite-sample Hoeffding alternative. Combined with family-wise error rate
control, StreamLTT yields a simultaneously certified policy region that
supports subsequent deployment selection. Across seven forecasting datasets
and seven deployment risks, StreamLTT places the certified frontier
substantially closer to the prescribed risk target than IID, robust-variance,
block-bootstrap, and concentration-based alternatives, while empirical
family-wise error closely follows the prescribed level. A finite-sample
ablation further shows that the predictable-risk formulation drives the
tighter risk frontier, while adaptation to predictable variance makes more
effective use of the available family-wise error budget. StreamLTT therefore
extends risk-certified policy selection to dependent deployment streams by
addressing both what risk should be certified and how it should be tested.
\end{abstract}

\input{Sec1_intro}

\input{Sec2_problem}
\input{Se3_method}
\input{Sec4_experiments}
\section{Conclusion}
\label{sec:conclusion}

We introduced StreamLTT for risk certified policy selection over dependent
and nonstationary deployment streams. StreamLTT replaces a fixed population
risk with average predictable risk, aligning certification with the
information available at each decision time and yielding martingale
fluctuations for policy level testing. This structure supports both a finite
sample bounded loss test and a variance adaptive test, which combine with
family wise error control to certify the candidate policy region. Across
seven forecasting datasets and seven deployment risks, StreamLTT-A produced
substantially tighter certified frontiers than IID, dependence aware,
bootstrap, and sequential concentration baselines, while empirical false
certification closely tracked the prescribed family wise level. The
finite sample ablation further separated the two sources of improvement:
the predictable risk formulation accounted for most of the frontier
tightening, while predictable variance adaptation primarily reduced testing
conservatism and made fuller use of the available family wise budget.
Together, these results extend LTT to evolving deployment streams by
addressing both what risk should be certified and how it should be tested
under dependence.

\clearpage

\section*{Reproducibility Statement}
Complete proofs and experimental details are provided in the appendix to
support reproduction of the theoretical and empirical results.

\section*{Ethics Statement}
This study used public datasets and needs no ethical approval.

\section*{AI Use Statement}
Generative AI was used only to assist with writing and language polishing.

\bibliographystyle{iclr2027_conference}
\bibliography{ref2}

\clearpage
\begin{center}
    {\Large\bf Supplementary File}
\end{center}

\appendix
\input{appendix}

\end{document}

%% file: math_commands.tex
\usepackage{amsmath,amsfonts,bm}

\def\eqref#1{equation~\ref{#1}}

\def\1{\bm{1}}

\DeclareMathAlphabet{\mathsfit}{\encodingdefault}{\sfdefault}{m}{sl}
\SetMathAlphabet{\mathsfit}{bold}{\encodingdefault}{\sfdefault}{bx}{n}

\newcommand{\E}{\mathbb{E}}



%% file: Sec1_intro.tex
\section{Introduction}
\label{sec:introduction}

Modern prediction systems are ultimately deployed through decisions rather
than raw predictions ~\citep{elmachtoub2022smart,wilder2019melding,donti2017task}. A practitioner may choose a confidence threshold,
prediction set width, abstention rule, alarm threshold, intervention rule, or
another operating policy subject to an explicit risk requirement
\citep{geifman2019selectivene,liu2019deepgamblers}. A broad literature has
developed statistical tools for uncertainty-aware and risk-controlled
decision-making, including selective classification, conformal prediction,
and conformal risk control
\citep{geifman2017selective,angelopoulos2021gentle,
angelopoulos2024crc}. Learn then Test (LTT) provides a natural framework for
policy selection: given a finite candidate family, it tests each policy
against a prescribed risk target and corrects for multiplicity so that the
resulting policy region is certified simultaneously
\citep{angelopoulos2021ltt}. For independent calibration losses, a common
implementation constructs $p$ values for individual policies using a central
limit theorem (CLT) approximation to empirical risk; we refer to this
instantiation as \emph{IID-LTT}. This separation between prediction,
certification, and subsequent selection makes LTT particularly attractive
when deployment must satisfy an explicit risk budget.

A central difficulty arises when certification data arrive as a stream.
Forecasts are temporally ordered, operating regimes evolve, and observations
may be coupled through past outcomes, recurring entities, or shared structure.
Even for a fixed policy, its conditional loss distribution can therefore
depend on the accumulated history and evolve throughout deployment, so a
single stationary marginal risk no longer describes the risks encountered
over time. This challenge has motivated a broad literature on \emph{online
calibration}, which uses sequential feedback to maintain predictive
calibration or coverage under distribution shift and nonexchangeability
\citep{gibbs2021aci,zaffran2022adaptive,gibbs2024conformal,
xu2023SPCI}.

Policy certification, however, poses a different statistical problem. In
LTT, the candidate family is fixed before certification, and each policy must
be tested against a risk constraint before multiplicity correction controls
the probability of certifying any unsafe candidate
\citep{angelopoulos2021ltt}. Under stream dependence, the challenge is
therefore not only how to construct a valid test, but also what risk should
be certified when the risk of a fixed policy can evolve along the stream.
Existing tools address the inference side of this problem: long-run variance
estimators adjust uncertainty for dependent averages
\citep{newey1987simple}, block resampling approximates sampling distributions
under local dependence \citep{politis1992circular}, and sequential
concentration controls deviations for adapted observations
\citep{howard2020timeuniform}. Yet these approaches still require a
pre-specified risk target, leaving open how risk itself should be defined
when it evolves with the observed history.

We address this gap by defining risk along the information flow of deployment.
Immediately before each outcome is revealed, we consider the conditional risk
of the corresponding decision given the available history and average these
risks along the stream, yielding the \emph{average predictable risk}. This
target does not require a stationary marginal distribution and reduces to the
standard population risk in the IID setting. Conditional centering further
turns empirical deviations from this target into a martingale sum, providing
the basis for \emph{StreamLTT}, our extension of LTT to policy certification
over dependent deployment streams. We develop a test that adapts to
predictable quadratic variation, together with a martingale Hoeffding
alternative~\citep{hoeffding1963probability} that requires only bounded losses
and provides finite-sample validity. The resulting $p$-values plug directly
into the multiple-testing layer of LTT~\citep{angelopoulos2021ltt}, yielding
a simultaneously certified policy region under family-wise error rate
control~\citep{holm1979simple}. A deployment policy can then be selected from
this region without repeating certification, retaining the modular selection
principle of LTT while aligning certification with the information structure
of an evolving stream.

Our experiments evaluate StreamLTT along two dimensions: the tightness of the
certified risk frontier and the control of false certification. Across seven
real-world forecasting datasets and seven deployment risks, StreamLTT
produces substantially tighter risk certificates than IID-LTT and
alternatives based on robust variance estimation, block resampling, and
concentration bounds, while empirical family-wise error closely follows the
prescribed level. A finite-sample ablation further separates the two sources
of improvement: the predictable-risk formulation already yields a tight
certified frontier, while adaptation to predictable variance makes more
effective use of the available family-wise error budget. These results
distinguish the benefit of changing the risk being certified from that of
adapting the statistical test to the variability of the stream.

Our contributions are threefold:
\begin{itemize}[leftmargin=*,itemsep=2pt,topsep=3pt]

    \item
    We formulate risk certification for policy selection under dependent and
    nonstationary deployment through \emph{average predictable risk}, which
    follows the information available at each decision time and recovers the
    standard population risk in the IID setting.

    \item
    We develop StreamLTT by using conditional centering to obtain martingale
    fluctuations, leading to a test that adapts to predictable variance and a
    Hoeffding alternative with finite sample validity. Combined with control
    of the family wise error rate, StreamLTT certifies an entire policy region
    while preserving validity under subsequent policy selection.

    \item
    Across seven forecasting datasets and seven deployment risks, StreamLTT
    yields substantially tighter certification than IID, robust variance,
    bootstrap, and concentration based alternatives while empirical family
    wise error follows the prescribed level. A finite sample ablation further
    separates the benefit of the predictable risk formulation from that of
    predictable variance adaptation.
\end{itemize}

%% file: Sec2_problem.tex
\section{Problem Formulation and Preliminaries}
\label{sec:prelim}

Consider a prediction system deployed over an ordered stream
$(X_1,Y_1),\ldots,(X_N,Y_N)$. At round $n$, the covariates $X_n$ and the
history from previous rounds are available before a deployment decision is
made, whereas the outcome $Y_n$ is revealed only afterward. Let
$\{\mathcal F_n\}_{n\ge0}$ denote the associated filtration, with the
history, current covariates, model output, and deployment decision at round
$n$ measurable with respect to $\mathcal F_{n-1}$. Here, we make no assumptions about independence nor
stationarity. The conditional law of
the stream may vary over time and depend on previously observed outcomes, recurring entities, or other shared context.

A finite family of deployment policies
$\Theta:=\{\theta_1,\ldots,\theta_M\}$ is fixed before certification. Each
policy may specify a confidence threshold, prediction-set width, alarm
threshold, or another operating rule applied to a frozen predictive model.
At round $n$, policy $\theta$ incurs an adapted bounded loss
$L_n(\theta)\in[0,B(\theta)]$, where the deterministic bound
$B(\theta)<\infty$ is fixed before certification. Given a prescribed risk
target $\alpha$ and family-wise error level $\delta$, the objective is to
construct a certified policy region such that the probability of including
any policy whose deployment risk exceeds $\alpha$ is controlled at level
$\delta$.

Learn-then-Test (LTT) casts risk-controlled policy selection as a
multiple-testing problem~\citep{angelopoulos2021ltt}. In the standard
setting, each candidate $\theta$ has population risk
$R(\theta):=\E[L(\theta)]$. A policy is considered safe when
$R(\theta)\le\alpha$, so LTT tests the complementary null
\begin{equation}
    H_j:
    R(\theta_j)>\alpha,
    \qquad
    j=1,\ldots,M.
    \label{eq:classical_ltt_null}
\end{equation}
A valid $p$-value is constructed for each candidate and passed to a
multiple-testing procedure. If the rejection rule controls the family-wise
error rate (FWER) at level $\delta$, the rejected hypotheses define a
certified region $\widehat\Theta_{\safe}$ satisfying
\begin{equation}
    \Prob\!\left(
        \widehat\Theta_{\safe}
        \subseteq
        \Theta_{\safe}
    \right)
    \ge
    1-\delta,
    \qquad
    \Theta_{\safe}
    :=
    \{\theta\in\Theta:R(\theta)\le\alpha\}.
    \label{eq:classical_ltt_guarantee}
\end{equation}
Thus, with probability at least $1-\delta$, every policy retained in the
certified region satisfies the prescribed risk requirement. Since the
guarantee holds simultaneously over the region, a final operating point may
subsequently be selected from $\widehat\Theta_{\safe}$ without retesting the
candidate family.

For independent calibration losses, LTT admits several policy level tests,
including finite sample bounded loss tests and asymptotic approximations.
In this work, we refer to the CLT based instantiation as \emph{IID-LTT}.
These standard constructions target a common population risk and rely on
independent calibration observations. They therefore do not directly address
the deployment setting considered here, where losses may be dependent and
their conditional risks may evolve with the observed history.

\paragraph{Problem formulation.}
Let $\mathcal R_N(\theta)$ denote the stream-level deployment risk associated
with policy $\theta$, whose precise form will be specified in
Section~\ref{sec:crc}. For a prescribed risk level $\alpha$, define the
corresponding safe region as
\begin{equation}
    \Theta_{\safe,N}
    :=
    \left\{
        \theta\in\Theta:
        \mathcal R_N(\theta)\le\alpha
    \right\}.
    \label{eq:stream_safe_region}
\end{equation}
Given the observed stream, the certification problem is to construct a
data-dependent region $\widehat\Theta_{\safe}\subseteq\Theta$ satisfying
\begin{equation}
    \Prob\!\left(
        \widehat\Theta_{\safe}
        \nsubseteq
        \Theta_{\safe,N}
    \right)
    =
    \Prob\!\left(
        \exists\,\theta\in\widehat\Theta_{\safe}:
        \mathcal R_N(\theta)>\alpha
    \right)
    \le\delta.
    \label{eq:stream_certification_goal}
\end{equation}
Thus, the statistical task is to specify a stream-level risk
$\mathcal R_N(\theta)$ compatible with the deployment information structure
and to construct valid policy-level tests for the unsafe hypotheses
$H_j:\mathcal R_N(\theta_j)>\alpha$. Section~\ref{sec:crc} instantiates
$\mathcal R_N$ as average predictable risk and develops the corresponding
dependence-aware tests.

%% file: Se3_method.tex
\section{StreamLTT: Learn-then-Test under Stream Dependence}
\label{sec:crc}

We now develop \emph{StreamLTT} for the stream-level certification problem
introduced in Section~\ref{sec:prelim}. Rather than imposing a common
population risk, StreamLTT averages the conditional risks encountered along
the deployment stream. Centering losses around this predictable target
induces martingale fluctuations, enabling finite-sample and variance-adaptive
policy tests under dependence. The resulting $p$-values are combined with
family-wise error control to obtain a simultaneously certified policy region.
\subsection{From Marginal to Predictable Risk}
\label{sec:predictable_risk}

For a fixed candidate policy $\theta\in\Theta$, let $L_n(\theta)$ denote its
bounded deployment loss at round $n$. Before $Y_n$ is revealed, define the
one-step predictable risk $
    p_n(\theta)
    :=
    \E[L_n(\theta)\mid\mathcal F_{n-1}],$
the conditional risk of deploying $\theta$ given the information available
when the decision is made. Unlike a fixed population risk, $p_n(\theta)$ may
change with the observed history and operating environment.

\begin{definition}[Average predictable risk]
\label{def:predictable_risk}
The average predictable risk of policy $\theta$ over the first $N$ rounds is
\begin{equation}
    R_{\mathrm{pred},N}(\theta)
    :=
    \frac1N\sum_{n=1}^{N}p_n(\theta).
    \label{eq:pred_risk}
\end{equation}
\end{definition}

The quantity $R_{\mathrm{pred},N}(\theta)$ therefore measures the average
conditional risk encountered along the realized deployment history, rather
than risk under a single stationary population. It allows the conditional
risk to vary across rounds and requires neither independence nor
stationarity. In the IID setting with common mean $R(\theta)$,
$p_n(\theta)=R(\theta)$ for every $n$, so
$R_{\mathrm{pred},N}(\theta)=R(\theta)$ and the standard LTT target is
recovered.

This target also yields the statistical structure needed for certification.
Let
\[
    \widehat R_N(\theta)
    :=
    \frac1N\sum_{n=1}^{N}L_n(\theta),
    \qquad
    Z_n(\theta)
    :=
    L_n(\theta)-p_n(\theta).
\]
Since
$\E[Z_n(\theta)\mid\mathcal F_{n-1}]=0$,
$\{Z_n(\theta)\}_{n\ge1}$ is a martingale-difference sequence, and hence
\begin{equation}
    \widehat R_N(\theta)-R_{\mathrm{pred},N}(\theta)
    =
    \frac1N\sum_{n=1}^{N}Z_n(\theta).
    \label{eq:risk_martingale_decomposition}
\end{equation}
Thus, dependence in the losses is retained in the risk target, while
fluctuations of empirical risk around that target admit martingale-based
inference.

\subsection{Stream-Aware Policy Testing}
\label{sec:statistical_certification}

For each candidate $\theta_j\in\Theta$, StreamLTT tests
$H_j:R_{\mathrm{pred},N}(\theta_j)>\alpha$, so rejecting $H_j$ certifies
$R_{\mathrm{pred},N}(\theta_j)\le\alpha$. By
Eq.~(\ref{eq:risk_martingale_decomposition}), empirical risk differs from
this target by a martingale average. This structure permits a finite-sample
test under bounded losses without specifying the dependence among the losses.

\begin{theorem}[Finite-sample StreamLTT test]
\label{thm:hoeffding_inid}
Suppose $L_n(\theta_j)$ is adapted to $\mathcal F_n$ and
$0\le L_n(\theta_j)\le B_j$, where $B_j:=B(\theta_j)$ is a deterministic
bound fixed before certification. Then
\begin{equation}
    p_j^{\mathrm H}
    :=
    \exp\!\left(
        -\frac{
            2N
            \bigl(
                \alpha-\widehat R_N(\theta_j)
            \bigr)_+^2
        }{
            B_j^2
        }
    \right)
    \label{eq:hoeffding_pvalue}
\end{equation}
is finite-sample super-uniform under
$H_j:R_{\mathrm{pred},N}(\theta_j)>\alpha$.
\end{theorem}

Theorem~\ref{thm:hoeffding_inid} provides a finite-sample guarantee using
only the deterministic loss bound $B_j$. The same robustness can make the
test conservative: its normalization reflects the worst-case loss range
rather than the conditional variability actually encountered along the
stream. When the latter is substantially smaller, a variance-sensitive
normalization can provide tighter certification. The proof is given in
Appendix~\ref{app:proof_hoeffding_inid}.

\textbf{Variance-adaptive testing.}
To capture this stream-specific variability, define the predictable
quadratic variation
\begin{equation}
    V_N^2(\theta)
    :=
    \sum_{n=1}^{N}
    \E\!\left[
        Z_n^2(\theta)
        \mid
        \mathcal F_{n-1}
    \right].
    \label{eq:predictable_variance}
\end{equation}
Whereas $B_j$ gives a uniform worst-case bound, $V_N^2(\theta)$ accumulates
the conditional variance of the martingale increments along the deployment
stream. We require this variation to admit a stable asymptotic scale.

\begin{assumption}[Predictable-variance stabilization]
\label{assum:lindeberg}
For every fixed $\theta_j\in\Theta$, there exists a deterministic sequence
$v_{N,j}^2>0$ such that $v_{N,j}^2\to\infty$ and
$V_N^2(\theta_j)/v_{N,j}^2\xrightarrow{p}1$.
\end{assumption}

Since bounded losses yield uniformly bounded martingale increments,
$v_{N,j}\to\infty$ makes the conditional Lindeberg condition automatic.
Together with Assumption~\ref{assum:lindeberg}, the martingale central limit
theorem then yields the variance-adaptive test below.

\begin{theorem}[Variance-adaptive StreamLTT test]
\label{thm:lf}
Suppose $L_n(\theta_j)$ is adapted to $\mathcal F_n$ and
$0\le L_n(\theta_j)\le B_j$, where $B_j:=B(\theta_j)$ is a deterministic
bound, and Assumption~\ref{assum:lindeberg} holds. Let
$\widehat V_N(\theta_j)$ consistently estimate the predictable variation,
such that
$\widehat V_N(\theta_j)/V_N(\theta_j)\xrightarrow{p}1$. Then
\begin{equation}
    p_j^{\mathrm{SL}}
    :=
    \Phi\!\left(
        \frac{
            N\bigl(
                \widehat R_N(\theta_j)-\alpha
            \bigr)
        }{
            \widehat V_N(\theta_j)
        }
    \right)
    \label{eq:pval_dep}
\end{equation}
is asymptotically super-uniform under
$H_j:R_{\mathrm{pred},N}(\theta_j)>\alpha$; namely, for every
$z\in[0,1]$,
\[
    \limsup_{N\to\infty}
    \Prob\!\left(
        p_j^{\mathrm{SL}}\le z
    \right)
    \le z.
\]
\end{theorem}

Theorem~\ref{thm:lf} trades the finite-sample robustness of
Theorem~\ref{thm:hoeffding_inid} for adaptation to the predictable
variability of the stream. When this variability is well below the
worst-case range, the adaptive normalization can reduce conservatism and
yield tighter certification. The proof is given in
Appendix~\ref{app:proof_lf}. Under IID losses, the construction recovers the
standard population-risk test.

\begin{corollary}[IID reduction]
\label{cor:iid_reduction}
Suppose $\{L_n(\theta_j)\}_{n=1}^{N}$ are IID with
$\E[L_n(\theta_j)]=R(\theta_j)$ and positive variance $\sigma_j^2$. Then
$R_{\mathrm{pred},N}(\theta_j)=R(\theta_j)$ and
$V_N^2(\theta_j)=N\sigma_j^2$. Consequently, the variance-adaptive
StreamLTT statistic reduces asymptotically to the standard IID
central-limit-theorem statistic for testing
$H_j:R(\theta_j)>\alpha$.
\end{corollary}

\textbf{Estimating the predictable variance.}
The adaptive test depends on $V_N(\theta)$, which is generally unknown.
For Bernoulli losses,
$\E[Z_n^2(\theta)\mid\mathcal F_{n-1}]
=p_n(\theta)(1-p_n(\theta))$.
Given an online estimate $\widehat p_n(\theta)$ of $p_n(\theta)$, we use
the plug-in estimator
\begin{equation}
    \widehat V_N^2(\theta)
    :=
    \sum_{n=1}^{N}
    \widehat p_n(\theta)
    \bigl(1-\widehat p_n(\theta)\bigr).
    \label{eq:vhat_plugin}
\end{equation}
The next result shows that consistency of the one-step risk estimates is
sufficient for the variance estimation required by
Theorem~\ref{thm:lf}.

\begin{lemma}[Consistency of predictable variance estimation]
\label{lem:variance_consistency}
Suppose
$N^{-1}\sum_{n=1}^{N}
(\widehat p_n(\theta)-p_n(\theta))^2\xrightarrow{p}0$
and
$V_N^2(\theta)/N\xrightarrow{p}\sigma_\theta^2>0$.
Then
$\widehat V_N(\theta)/V_N(\theta)\xrightarrow{p}1$.
Moreover, Assumption~\ref{assum:lindeberg} holds with
$v_N^2=N\sigma_\theta^2$.
\end{lemma}

Lemma~\ref{lem:variance_consistency} therefore links estimation of the
one-step predictable risks directly to the normalization used by the
variance-adaptive test. The proof is given in
Appendix~\ref{app:variance_estimation}.
\subsection{From Policy Tests to a Certified Region}
\label{sec:certified_region}

The preceding results establish valid policy-level tests for predictable
risk. We now lift these guarantees from individual candidates to the entire
policy family. Let $\mathcal R_N\subseteq\{1,\ldots,M\}$ denote the rejection
set returned by the multiple-testing procedure and define
$\widehat\Theta_{\safe}:=\{\theta_j:j\in\mathcal R_N\}$. The corresponding
safe region is
$\Theta_{\safe,N}:=\{\theta\in\Theta:
R_{\mathrm{pred},N}(\theta)\le\alpha\}$.
Controlling false rejections across the candidate family then yields
simultaneous certification of $\widehat\Theta_{\safe}$.

\begin{theorem}[Simultaneous StreamLTT certification]
\label{thm:certified_region}
Let $\widehat\Theta_{\safe}$ be obtained by applying FWER control at level
$\delta$ to the finite-sample StreamLTT $p$-values in
Theorem~\ref{thm:hoeffding_inid}. Then, for any possibly data-dependent
selection $\widehat\theta\in\widehat\Theta_{\safe}$ whenever
$\widehat\Theta_{\safe}\neq\emptyset$,
\begin{equation}
\begin{aligned}
    \Prob\!\left(
        \widehat\Theta_{\safe}\subseteq\Theta_{\safe,N}
    \right)
    &\ge 1-\delta,
    &
    \Prob\!\left(
        \widehat\Theta_{\safe}\neq\emptyset,\,
        R_{\mathrm{pred},N}(\widehat\theta)>\alpha
    \right)
    &\le\delta.
\end{aligned}
\label{eq:ltt_guarantee}
\end{equation}

For a fixed finite candidate family, applying FWER control at level
$\delta$ to the variance-adaptive $p$-values in
Theorem~\ref{thm:lf} yields
\begin{equation}
\begin{aligned}
    \liminf_{N\to\infty}
    \Prob\!\left(
        \widehat\Theta_{\safe}\subseteq\Theta_{\safe,N}
    \right)
    &\ge 1-\delta,
    &
    \limsup_{N\to\infty}
    \Prob\!\left(
        \widehat\Theta_{\safe}\neq\emptyset,\,
        R_{\mathrm{pred},N}(\widehat\theta)>\alpha
    \right)
    &\le\delta.
\end{aligned}
\label{eq:ltt_guarantee_asymptotic}
\end{equation}
\end{theorem}

Theorem~\ref{thm:certified_region} lifts the policy-level validity established
above to simultaneous certification of the candidate family. The
policy-level tests account for dependence along the deployment stream,
whereas FWER control accounts for multiplicity across candidate policies.
The first guarantee ensures that the entire certified region is safe with
probability at least $1-\delta$; the second shows that any subsequent
data-dependent selection from this region inherits the same risk control
without additional testing.

\textbf{Complete procedure.}
Algorithm~\ref{alg:streamltt} summarizes the resulting StreamLTT pipeline.
Each candidate is evaluated using either the finite-sample Hoeffding test or
the variance-adaptive test, and the resulting $p$-values are combined with
FWER control to form $\widehat\Theta_{\safe}$. A deployment policy may then
be selected from the certified region according to the downstream objective.
For $M$ candidates, StreamLTT maintains $\mathcal O(M)$ running summaries and
requires $\mathcal O(M)$ computation per stream round.

\begin{algorithm}[t]
\caption{StreamLTT for risk-certified policy selection}
\label{alg:streamltt}
\begin{algorithmic}[1]
\REQUIRE Certification stream $\{(X_n,Y_n)\}_{n=1}^{N}$; candidate family
$\Theta=\{\theta_1,\ldots,\theta_M\}$; loss $L_n(\theta)$; risk target
$\alpha$; bounds $\{B(\theta_j)\}_{j=1}^{M}$; FWER level $\delta$;
$\textsc{Mode}\in\{\textsc{Adaptive},\textsc{Hoeffding}\}$.
\ENSURE Certified region $\widehat\Theta_{\safe}$; optional policy
$\widehat\theta$.

\STATE Freeze the candidate family and all test-specific quantities.
\STATE Initialize empirical-risk summaries and, for $\textsc{Adaptive}$ mode,
predictable-variance summaries.

\FOR{$n=1,\ldots,N$}
    \STATE Observe $\mathcal F_{n-1}$ and form decisions for all
    $\theta_j\in\Theta$.
    \STATE Reveal $Y_n$ and record $L_n(\theta_j)$ for all
    $\theta_j\in\Theta$.
    \IF{$\textsc{Mode}=\textsc{Adaptive}$}
        \STATE Estimate $\widehat p_n(\theta_j)$ and update the
        predictable-variance summaries via Eq.~(\ref{eq:vhat_plugin}).
    \ENDIF
\ENDFOR

\FOR{$j=1,\ldots,M$}
    \STATE Compute
    $\widehat R_N(\theta_j)\leftarrow
    N^{-1}\sum_{n=1}^{N}L_n(\theta_j)$.
    \IF{$\textsc{Mode}=\textsc{Adaptive}$}
        \STATE Compute $p_j\leftarrow p_j^{\mathrm{SL}}$ via
        Eq.~(\ref{eq:pval_dep}).
    \ELSE
        \STATE Compute $p_j\leftarrow p_j^{\mathrm H}$ via
        Eq.~(\ref{eq:hoeffding_pvalue}).
    \ENDIF
\ENDFOR

\STATE Apply FWER control at level $\delta$ to $\{p_j\}_{j=1}^{M}$ and form
$\widehat\Theta_{\safe}$ from the rejected hypotheses.
\IF{$\widehat\Theta_{\safe}=\emptyset$}
    \STATE \textbf{return} ``no certified policy''.
\ENDIF
\STATE Optionally select $\widehat\theta\in\widehat\Theta_{\safe}$ according
to the deployment criterion.
\STATE \textbf{return} $\widehat\Theta_{\safe}$ and optional
$\widehat\theta$.
\end{algorithmic}
\end{algorithm}

%% file: Sec4_experiments.tex
\section{Experiments}
\label{sec:experiments}

We evaluate StreamLTT along two complementary dimensions of risk
certification: the tightness of the certified risk frontier and the control
of false certification. We refer to the variance adaptive test in
Theorem~\ref{thm:lf} as \emph{StreamLTT-A} and the finite sample Hoeffding
test in Theorem~\ref{thm:hoeffding_inid} as \emph{StreamLTT-H}. Our primary
comparison evaluates StreamLTT-A against IID, dependence aware, and
sequential certification baselines. StreamLTT-H provides a finite sample
counterpart that isolates the effect of predictable variance adaptation.

\subsection{Experimental Setup}
\label{sec:experimental_setup}

\textbf{Datasets and forecasting protocol.}
We consider seven real world forecasting datasets spanning three domains:
Electricity~\citep{trindade2015electricity}, Appliances
Energy~\citep{candanedo2017data}, and Household
Power~\citep{hebrail2006household} for energy forecasting; PEMS04 and
PEMS08~\citep{guo2019attention} for traffic forecasting; and Beijing
Weather~\citep{chen2017beijing} and Jena
Weather~\citep{jena_climate} for temperature forecasting. The datasets vary
in sequence length and sampling frequency, and their temporal order is
preserved throughout the experiment. Each dataset is split chronologically
into a training set (20\%), a validation set (2\%), a calibration set (8\%),
and a certification stream (70\%). A temporal convolutional network (TCN) \citep{bai2018tcn}is
trained for one step forecasting on the training set, selected on the
validation set, and then frozen. The calibration set determines auxiliary
quantities, such as severe miss tolerances and regime thresholds, which are
fixed before certification. The remaining observations are processed
sequentially as the certification stream, with the same frozen forecasts
shared across all certification methods. At each round, only information
available before the corresponding outcome is revealed is used; history
dependent decisions may additionally use previously revealed outcomes.
Model specifications, preprocessing, and further protocol details are
provided in Appendix~\ref{app:forecasting_protocol}.

\textbf{Deployment risks.}
We consider seven binary deployment risks spanning pointwise, regime
dependent, directional, and history dependent errors, as summarized in
Table~\ref{tab:risk_definitions}. These risks range from standard forecasting
failures to decisions whose losses depend on the current operating regime or
previously observed outcomes, providing diverse dependence structures for
stream level certification. All auxiliary quantities are fixed before
certification, and each risk defines a separate certification problem.
Complete risk constructions are provided in
Appendix~\ref{app:risk_construction}.

\begin{table}[!t]
\centering
\caption{
Binary deployment risks used for certification. All auxiliary thresholds are
determined from tuning data and fixed before certification.
}
\label{tab:risk_definitions}
\small
\setlength{\tabcolsep}{5pt}
\renewcommand{\arraystretch}{1.08}
\begin{tabular}{lllc}
\toprule
Risk & Structure & Error event & $\alpha$ \\
\midrule
Interval miss
& Pointwise
& $Y_i\notin C_m(X_i)$
& .10 \\
Lower-tail miss
& Pointwise
& $Y_i<\ell_i(m)$
& .10 \\
Severe miss
& Pointwise
& $|Y_i-\widehat\mu_i|>ms_i+e_0$
& .10 \\
Task-regime miss
& Regime
& $G_i=1,\;Y_i\notin C_m(X_i)$
& .05 \\
Opposite direction
& Temporal
& $a_i(\theta)z_i^d=-1$
& .10 \\
Missed reversal
& History
& $r_i=1,\;\widehat r_i(\theta)=0$
& .10 \\
False reversal
& History
& $r_i=0,\;\widehat r_i(\theta)=1$
& .10 \\
\bottomrule
\end{tabular}
\end{table}

\textbf{Methods.}
Our primary method is \emph{StreamLTT-A}, the variance adaptive test in
Theorem~\ref{thm:lf}. We compare it with \emph{IID-LTT}
\citep{angelopoulos2021ltt}, \emph{LTT-HB-Holm}
\citep{hoeffding1963probability,bentkus2004hoeffding},
\emph{HAC-CLT-Holm}~\citep{newey1987simple},
\emph{CircularBlockBootstrap-Holm}~\citep{politis1992circular}, and
\emph{EB-Howard-Bonferroni}~\citep{howard2020timeuniform}, covering IID
inference, bounded loss concentration, long run variance correction, block
resampling, and sequential concentration, respectively. We also evaluate
\emph{StreamLTT-H}, the finite sample test in
Theorem~\ref{thm:hoeffding_inid}, which retains the same predictable risk
target as StreamLTT-A while removing predictable variance adaptation. All
methods use the same risk targets and family wise levels. Further details are
provided in Appendix~\ref{app:certification_methods}.

\textbf{Evaluation protocol and metrics.}
We evaluate all $7\times7=49$ dataset--risk combinations at
$\delta\in\{0.10,0.20\}$ using $2{,}000$ independent repetitions per
configuration. We assess certification through risk tightness and false
certification control. Empirical FWER is the fraction of repetitions in
which at least one certified policy has predictable risk above $\alpha$.
For each nonempty certified region, $R_{\max}$ denotes its largest
predictable risk, and $\overline R_{\max}$ its mean across repetitions.
We report the risk gap $\alpha-\overline R_{\max}$ and FWER gap
$\delta-\widehat{\mathrm{FWER}}$, where values near zero indicate fuller use
of the respective risk and family wise budgets. We additionally report the
certified fraction, the proportion of candidate policies retained after
certification. Further evaluation details, risk wise and dataset level
results, and Monte Carlo uncertainty are provided in
Appendix~\ref{app:evaluation_protocol}.
\subsection{Results}
\label{sec:results}

\begin{table*}[t]
\centering
\caption{
Representative certification results at $\delta=0.10$ across seven
datasets. Each entry reports
$\overline R_{\max}/\widehat{\mathrm{FWER}}$.
The three risks represent two sided, one sided, and history dependent
deployment errors.
}
\label{tab:main_results}
\footnotesize
\setlength{\tabcolsep}{2.2pt}
\renewcommand{\arraystretch}{1.06}
\begin{tabular}{llccccccc}
\toprule
Risk & Method
& Electricity & Appliances & Household
& PEMS04 & PEMS08 & Beijing & Jena \\
\midrule

\multirow{6}{*}{\shortstack[l]{Interval\\($\alpha=.10$)}}
& IID-LTT
& .051/.028 & .046/.013 & .048/.020 & .051/.017
& .044/.015 & .047/.019 & .045/.012 \\
& LTT-HB-Holm
& .050/.006 & .045/.004 & .047/.009 & .051/.006
& .043/.006 & .046/.005 & .045/.004 \\
& HAC-CLT-Holm
& .051/.035 & .046/.014 & .048/.025 & .051/.021
& .044/.019 & .047/.024 & .045/.017 \\
& BlockBootstrap-Holm
& .051/.034 & .046/.015 & .048/.025 & .052/.021
& .044/.019 & .047/.023 & .045/.018 \\
& EB-Howard-Bonf.
& .050/.000 & .045/.000 & .047/.000 & .050/.000
& .043/.000 & .046/.000 & .044/.000 \\
& \textbf{StreamLTT-A}
& \textbf{.096/.084} & \textbf{.096/.082} & \textbf{.097/.102}
& \textbf{.096/.094} & \textbf{.095/.087} & \textbf{.096/.089}
& \textbf{.096/.083} \\
\midrule

\multirow{6}{*}{\shortstack[l]{Lower tail\\($\alpha=.10$)}}
& IID-LTT
& .055/.045 & .069/.031 & .053/.032 & .065/.025
& .066/.026 & .069/.020 & .064/.025 \\
& LTT-HB-Holm
& .053/.014 & .069/.010 & .052/.009 & .065/.009
& .066/.009 & .069/.004 & .063/.007 \\
& HAC-CLT-Holm
& .055/.052 & .069/.037 & .053/.037 & .066/.030
& .066/.031 & .070/.023 & .064/.032 \\
& BlockBootstrap-Holm
& .055/.052 & .069/.036 & .053/.036 & .066/.031
& .066/.031 & .070/.023 & .064/.031 \\
& EB-Howard-Bonf.
& .052/.000 & .068/.000 & .052/.000 & .065/.000
& .065/.000 & .069/.000 & .063/.000 \\
& \textbf{StreamLTT-A}
& \textbf{.097/.095} & \textbf{.096/.097} & \textbf{.095/.093}
& \textbf{.097/.080} & \textbf{.097/.088} & \textbf{.096/.081}
& \textbf{.097/.085} \\
\midrule

\multirow{6}{*}{\shortstack[l]{Missed reversal\\($\alpha=.10$)}}
& IID-LTT
& .080/.087 & .064/.026 & .075/.040 & .048/.016
& .050/.018 & .089/.091 & .081/.047 \\
& LTT-HB-Holm
& .077/.030 & .063/.008 & .074/.015 & .048/.005
& .049/.006 & .086/.031 & .080/.018 \\
& HAC-CLT-Holm
& .081/.099 & .064/.033 & .075/.045 & .049/.022
& .050/.023 & .090/.099 & .081/.052 \\
& BlockBootstrap-Holm
& .081/.097 & .064/.033 & .075/.047 & .049/.022
& .050/.023 & .090/.099 & .081/.053 \\
& EB-Howard-Bonf.
& .073/.000 & .063/.000 & .074/.000 & .048/.001
& .049/.000 & .079/.000 & .080/.000 \\
& \textbf{StreamLTT-A}
& \textbf{.097/.088} & \textbf{.095/.092} & \textbf{.095/.085}
& \textbf{.095/.090} & \textbf{.095/.089} & \textbf{.096/.093}
& \textbf{.082/.092} \\
\bottomrule
\end{tabular}
\end{table*}

\begin{figure*}[t]
    \centering
    \includegraphics[width=\textwidth]
    {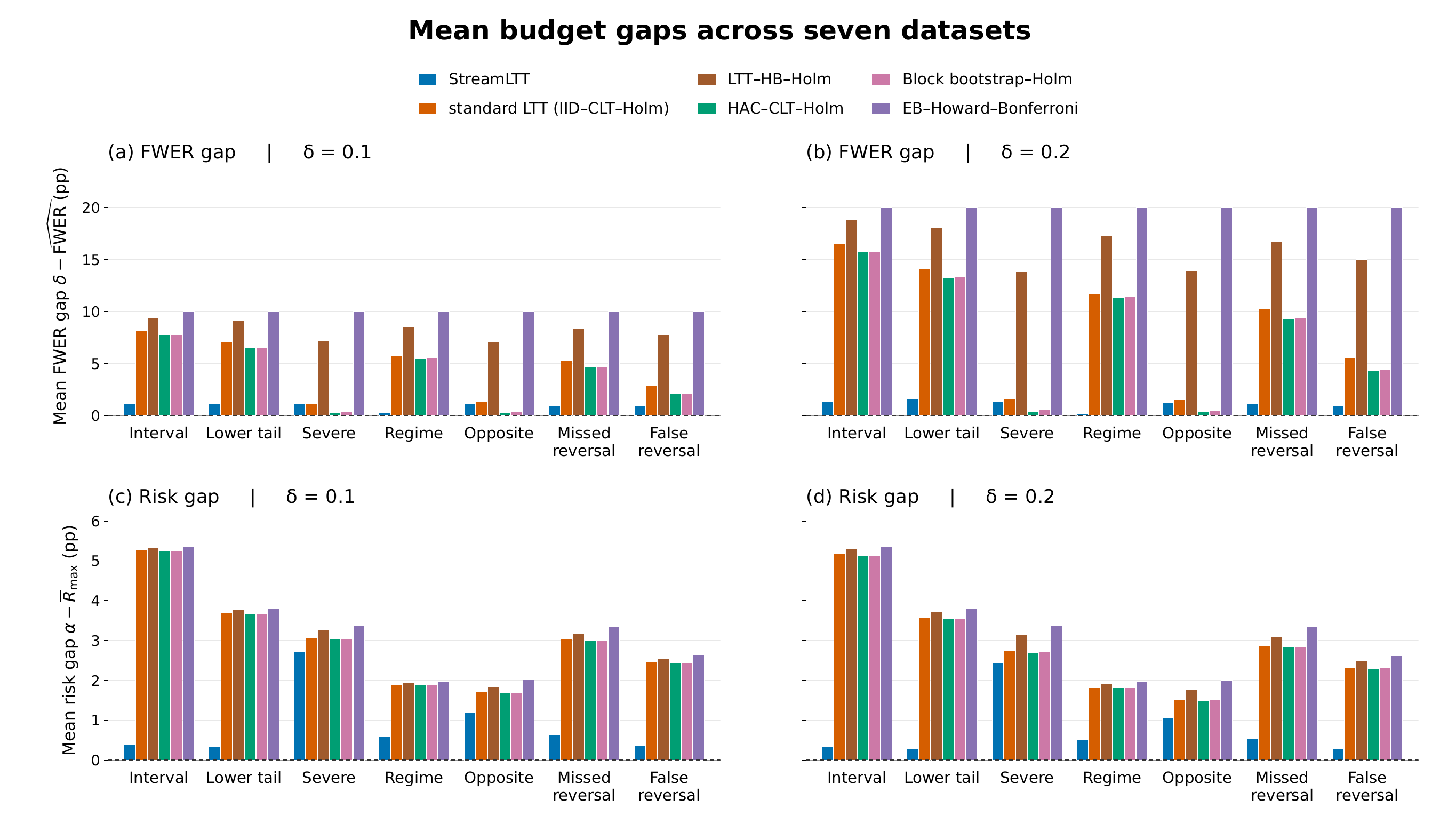}
    \vspace*{-0.5cm}
    \caption{
    Mean family wise error rate (FWER) and risk gaps across seven deployment
    risks. The top row reports
    $\delta-\widehat{\mathrm{FWER}}$ at $\delta=0.10$ and $0.20$; the
    bottom row reports
    $\alpha-\overline R_{\max}$ at the corresponding family wise levels.
    Each bar averages over the seven datasets. Values closer to zero indicate
    tighter utilization of the corresponding certification budget.
    }
    \label{fig:budget_gaps}
\end{figure*}

\textbf{StreamLTT-A approaches both certification boundaries.}
Figure~\ref{fig:budget_gaps} shows that StreamLTT-A uses substantially more
of both the risk and family wise error budgets. At $\delta=0.10$, its mean
risk gap is $0.0090$, placing the certified frontier close to the prescribed
risk boundary, compared with $0.0302$ for IID-LTT and approximately
$0.030$--$0.032$ for the dependence and concentration based alternatives.
This tighter frontier is not accompanied by excessive false certification:
StreamLTT-A attains empirical FWER $0.0902$, close to the prescribed
$\delta=0.10$. In contrast, the competing procedures leave considerably
more slack in the risk budget, the family wise budget, or both. StreamLTT-A
therefore expands certification toward the admissible risk boundary while
maintaining the intended family wise control.

Table~\ref{tab:main_results} resolves this aggregate improvement at the
dataset level for three representative risks; complete results for all seven
risks are provided in the appendix. For interval miss, the baseline
frontiers remain between $0.043$ and $0.052$, whereas StreamLTT-A reaches
$0.095$--$0.097$ across all seven datasets. The separation persists for
lower tail miss, where the baseline values range from $0.052$ to $0.070$
while StreamLTT-A remains at $0.095$--$0.097$. For the history dependent
missed reversal risk, the gain is particularly pronounced on PEMS04 and
PEMS08, where StreamLTT-A reaches $0.095$ compared with approximately
$0.048$--$0.050$ for the competing methods. Across all 49 dataset--risk
combinations, StreamLTT-A retains $76.7\%$ of the candidate family while
maintaining mean empirical FWER $0.0902$ at the prescribed
$\delta=0.10$.

\textbf{Dependence aware inference alone does not remove certification
slack.}
The risk gaps help isolate the source of the improvement. At
$\delta=0.10$, IID-LTT has a mean gap of $0.0302$, while HAC-CLT-Holm and
CircularBlockBootstrap-Holm attain $0.0300$ despite explicitly accounting
for temporal dependence. StreamLTT-A reduces this gap to $0.0090$.
Thus, adjusting the sampling distribution for dependence has little effect
on the certified frontier when the underlying risk target remains unchanged.
StreamLTT instead certifies average predictable risk, conditioning risk on
the information available at each decision time. The empirical separation
therefore supports the distinction in Section~\ref{sec:predictable_risk}:
dependence aware inference changes how risk is tested, whereas StreamLTT
also changes what risk is certified along the deployment stream.

\textbf{StreamLTT-A adapts to the available family-wise budget.}
Increasing $\delta$ from $0.10$ to $0.20$ provides a direct test of whether
the certification procedure responds to a larger tolerance for false
certification. StreamLTT-A's empirical FWER increases from $0.0902$ to
$0.1888$, closely tracking the corresponding prescribed levels, while its
mean risk gap remains small and decreases from $0.0090$ to $0.0078$.
The baselines respond much less strongly: at $\delta=0.20$, IID-LTT,
HAC-CLT-Holm, and CircularBlockBootstrap-Holm reach FWER only $0.1125$,
$0.1217$, and $0.1208$, respectively, while LTT-HB-Holm remains at $0.0376$
and EB-Howard-Bonferroni near zero. Their risk gaps also remain between
$0.0283$ and $0.0321$. Thus, increasing the family wise budget primarily
expands certification under StreamLTT-A, rather than leaving the additional
error budget unused.

\textbf{Ablation: predictable variance adaptation primarily improves
family wise budget utilization.}
StreamLTT-H and StreamLTT-A certify the same average predictable risk and
differ only in the policy level test, allowing us to isolate the effect of
predictable variance adaptation. Their certified frontiers are already
similar: at $\delta=0.10$ and $0.20$, StreamLTT-H attains mean risk gaps
$0.0118$ and $0.0112$, compared with $0.0090$ and $0.0078$ for
StreamLTT-A. The certified fractions are likewise close, increasing from
$75.3\%$ to $76.7\%$ at $\delta=0.10$ and from $75.7\%$ to $77.1\%$ at
$\delta=0.20$.

The main difference appears in false certification. StreamLTT-H has empirical
FWER only $0.0001$ and $0.0006$ at the two family wise levels, leaving
nearly the entire error budget unused, whereas StreamLTT-A reaches $0.0902$
and $0.1888$. Thus, the tight frontier is already largely present when
predictable risk is paired with the finite sample test; predictable variance
adaptation provides a modest additional improvement in frontier tightness
and certified fraction, but its dominant empirical effect is to reduce the
conservatism of policy level testing and make substantially fuller use of
the available family wise budget.

%% file: appendix.tex
\input{related_works}
\input{appendix_proofs.tex}
\input{Expeiments_detail}
\input{Additional_Exp}

%% file: related_works.tex
\section{Related Works}
\label{app:related_work}

StreamLTT is related to the following works on risk controlled prediction, sequential calibration,
inference under dependence, and multiple testing. 

\subsection{Risk Control and Sequential Calibration}

Conformal prediction provides distribution free predictive uncertainty under
exchangeability
\citep{vovk2005conformal,shafer2008tutorial,lei2018distribution,
angelopoulos2021gentle}, with extensions to regression, classification, and
more general predictive settings
\citep{romano2020aps}. Related work on
selective prediction studies decisions with abstention and the tradeoff
between prediction frequency and reliability
\citep{chow1970optimum,elyaniv2010foundations,geifman2017selective,
geifman2019selectivene}. More recently, conformal risk control extends
calibration from coverage to general monotone losses
\citep{angelopoulos2024crc}, while Learn then Test (LTT) casts risk
controlled policy selection as a multiple testing problem
\citep{angelopoulos2021ltt}. StreamLTT retains the LTT architecture of
testing a fixed candidate family and certifying the rejected policies
simultaneously, but replaces the stationary population risk with average
predictable risk.

A complementary literature studies conformal calibration when data arrive
sequentially or exchangeability fails. Adaptive Conformal Inference and its
extensions use sequential feedback to maintain coverage under distribution
shift
\citep{gibbs2021aci,zaffran2022adaptive,gibbs2024conformal}, while SPCI
exploits temporal structure in sequential conformity scores
\citep{xu2023SPCI}. These methods primarily adapt predictive calibration as
new observations arrive. StreamLTT instead keeps the candidate policy family
fixed and asks which policies satisfy a deployment risk constraint. Online
calibration and StreamLTT therefore operate at complementary stages:
prediction can adapt over time, while certification evaluates a fixed family
of deployment policies.

\subsection{Inference under Dependence}

Classical inference under temporal dependence often retains the target
estimand while modifying its sampling approximation. HAC methods estimate
long run variance under serial dependence
\citep{newey1987simple,andrews1991heteroskedasticity}, while block bootstrap
methods preserve dependence by resampling consecutive observations
\citep{kunsch1989jackknife,politis1992circular,politis1994stationary,
lahiri2003resampling}. These approaches motivate our HAC and block bootstrap
baselines. They address how an empirical statistic should be normalized or
resampled under dependence, but do not by themselves specify which deployment
risk should replace a stationary marginal expectation when conditional risk
changes along the stream.

Martingale concentration provides a complementary route to sequential
inference. Classical bounded increment and variance sensitive inequalities
include Hoeffding, Bentkus, and Freedman type bounds
\citep{hoeffding1963probability,bentkus2004hoeffding,freedman1975tail},
with self normalized extensions developed for martingale processes
\citep{delapena2009selfnormalized}. More recent work uses nonnegative
supermartingales to obtain time uniform concentration and confidence
sequences
\citep{howard2020timeuniform,howard2021timeuniform}, including betting based
inference for bounded means
\citep{waudbysmith2024estimating}. StreamLTT uses this martingale perspective
after defining the risk target: centering each loss by its one step
predictable risk yields a martingale difference sequence. Boundedness then
gives the finite sample test, while predictable quadratic variation supports
the variance adaptive construction.

\subsection{Simultaneous Policy Certification}

Certification across a finite policy family additionally requires
multiplicity control. Classical procedures such as Bonferroni and Holm
provide control of the family wise error rate, with related stepwise and
closed testing procedures providing further refinements
\citep{holm1979simple,marcus1976closed,hochberg1988sharper}. LTT connects
these tools to policy selection by interpreting rejection of an unsafe risk
null as certification of the corresponding policy
\citep{angelopoulos2021ltt}. StreamLTT preserves this multiple testing layer:
once valid $p$ values for predictable risk are available, standard family
wise correction yields a simultaneously certified region and permits
subsequent selection within that region.


%% file: appendix_proofs.tex
\section{Proofs for StreamLTT}
\label{app:streamltt_proofs}

This appendix provides the proofs and supporting arguments for
Section~\ref{sec:crc}. We follow the order of the main text: the
finite-sample StreamLTT test, the variance-adaptive StreamLTT test,
consistency of the predictable-variance estimator, and simultaneous
certification with post-certification selection.

\subsection{Proof of the Finite-Sample StreamLTT Test}
\label{app:proof_hoeffding_inid}

\begin{proof}[Proof of Theorem~\ref{thm:hoeffding_inid}]
Fix $\theta_j\in\Theta$ and suppress the argument $\theta_j$. Recall
$p_n=\E[L_n\mid\mathcal F_{n-1}]$ and $Z_n=L_n-p_n$. By construction,
$\E[Z_n\mid\mathcal F_{n-1}]=0$, so $\{Z_n\}_{n\ge1}$ is a
martingale-difference sequence with respect to
$\{\mathcal F_n\}_{n\ge0}$.

Since $0\le L_n\le B_j$ and conditional expectation preserves these bounds,
$0\le p_n\le B_j$. Conditional on $\mathcal F_{n-1}$,
\[
    -p_n\le Z_n\le B_j-p_n,
\]
so $Z_n$ lies in an interval of length $B_j$. Hoeffding's lemma therefore
gives, for every $s>0$,
\[
    \E\!\left[
        e^{-sZ_n}
        \mid
        \mathcal F_{n-1}
    \right]
    \le
    \exp\!\left(
        \frac{s^2B_j^2}{8}
    \right).
\]

Define
\[
    M_k
    :=
    \exp\!\left(
        -s\sum_{n=1}^{k}Z_n
        -\frac{k s^2B_j^2}{8}
    \right),
    \qquad
    M_0:=1.
\]
Then $\{M_k,\mathcal F_k\}_{k=0}^{N}$ is a nonnegative supermartingale,
since
\[
\begin{aligned}
    \E[M_k\mid\mathcal F_{k-1}]
    &=
    M_{k-1}
    e^{-s^2B_j^2/8}
    \E\!\left[
        e^{-sZ_k}
        \mid
        \mathcal F_{k-1}
    \right] \le M_{k-1}.
\end{aligned}
\]
Hence $\E[M_N]\le1$.

For any $t>0$, on the event
$\{\sum_{n=1}^{N}Z_n\le-Nt\}$,
\[
    M_N
    \ge
    \exp\!\left(
        sNt-\frac{Ns^2B_j^2}{8}
    \right).
\]
Markov's inequality therefore yields
\[
    \Prob\!\left(
        \sum_{n=1}^{N}Z_n\le-Nt
    \right)
    \le
    \exp\!\left(
        -sNt+\frac{Ns^2B_j^2}{8}
    \right).
\]
Optimizing over $s>0$ at $s=4t/B_j^2$ gives
\begin{equation}
    \Prob\!\left(
        \widehat R_N-R_{\mathrm{pred},N}\le-t
    \right)
    \le
    \exp\!\left(
        -\frac{2Nt^2}{B_j^2}
    \right),
    \label{eq:app_martingale_hoeffding}
\end{equation}
where Eq.~(\ref{eq:risk_martingale_decomposition}) was used.

Fix $z\in(0,1)$ and define
$t_z:=B_j\sqrt{\log(1/z)/(2N)}$. By
Eq.~(\ref{eq:hoeffding_pvalue}),
$p_j^{\mathrm H}\le z$ implies
$\widehat R_N\le\alpha-t_z$. Under
$H_j:R_{\mathrm{pred},N}>\alpha$,
\[
    \widehat R_N-R_{\mathrm{pred},N}
    <
    -t_z.
\]
Thus
\[
    \{p_j^{\mathrm H}\le z\}
    \subseteq
    \{
        \widehat R_N-R_{\mathrm{pred},N}\le-t_z
    \}.
\]
Applying Eq.~(\ref{eq:app_martingale_hoeffding}),
\[
\begin{aligned}
    \Prob_{H_j}\!\left(
        p_j^{\mathrm H}\le z
    \right)
    &\le
    \exp\!\left(
        -\frac{2Nt_z^2}{B_j^2}
    \right) = z.
\end{aligned}
\]
The cases $z=0$ and $z=1$ are immediate. Hence
$p_j^{\mathrm H}$ is finite-sample super-uniform under
$H_j:R_{\mathrm{pred},N}(\theta_j)>\alpha$.
\end{proof}

\subsection{Proof of the Variance-Adaptive StreamLTT Test}
\label{app:proof_lf}

\begin{proof}[Proof of Theorem~\ref{thm:lf}]
Fix $\theta_j\in\Theta$ and suppress the argument $\theta_j$. Recall
$p_n=\E[L_n\mid\mathcal F_{n-1}]$ and $Z_n=L_n-p_n$. By construction,
$\{Z_n\}_{n\ge1}$ is a martingale-difference sequence with respect to
$\{\mathcal F_n\}_{n\ge0}$, and
Eq.~(\ref{eq:risk_martingale_decomposition}) gives
\[
    N\bigl(
        \widehat R_N-R_{\mathrm{pred},N}
    \bigr)
    =
    \sum_{n=1}^{N}Z_n.
\]

Since $0\le L_n\le B_j$, we have $0\le p_n\le B_j$ and therefore
$|Z_n|\le B_j$. By Assumption~\ref{assum:lindeberg}, there exists a
deterministic sequence $v_{N,j}^2>0$ such that
$v_{N,j}^2\to\infty$ and
\[
    \frac{V_N^2}{v_{N,j}^2}
    \xrightarrow{p}1,
\]
where $V_N^2$ is defined in Eq.~\ref{eq:predictable_variance}.

For any $\varepsilon>0$, $v_{N,j}\to\infty$ implies
$\varepsilon v_{N,j}>B_j$ for all sufficiently large $N$. Hence
\[
    \frac1{v_{N,j}^2}
    \sum_{n=1}^{N}
    \E\!\left[
        Z_n^2
        \ind\{|Z_n|>\varepsilon v_{N,j}\}
        \mid
        \mathcal F_{n-1}
    \right]
    =0
\]
eventually. Thus the conditional Lindeberg condition holds.

Combining the conditional Lindeberg condition with predictable-variance
stabilization, the martingale central limit theorem gives
\[
    \frac{
        \sum_{n=1}^{N}Z_n
    }{
        v_{N,j}
    }
    \Rightarrow
    \mathcal N(0,1).
\]
Moreover,
$V_N/v_{N,j}\xrightarrow{p}1$, so Slutsky's theorem yields
\[
    \frac{
        \sum_{n=1}^{N}Z_n
    }{
        V_N
    }
    \Rightarrow
    \mathcal N(0,1).
\]

By the variance-consistency condition in
Theorem~\ref{thm:lf},
$\widehat V_N/V_N\xrightarrow{p}1$. Applying Slutsky's theorem again,
\begin{equation}
    T_N
    :=
    \frac{
        N\bigl(
            \widehat R_N-R_{\mathrm{pred},N}
        \bigr)
    }{
        \widehat V_N
    }
    \Rightarrow
    \mathcal N(0,1).
    \label{eq:app_studentized_clt}
\end{equation}

The statistic underlying Eq.~(\ref{eq:pval_dep}) can be written as
\[
\begin{aligned}
    \frac{
        N(\widehat R_N-\alpha)
    }{
        \widehat V_N
    }
    &=
    T_N
    +
    \frac{
        N(R_{\mathrm{pred},N}-\alpha)
    }{
        \widehat V_N
    }.
\end{aligned}
\]
Under $H_j:R_{\mathrm{pred},N}>\alpha$, the second term is positive.
Since $\Phi$ is increasing,
\[
    p_j^{\mathrm{SL}}
    \ge
    \Phi(T_N).
\]
Consequently, for every $z\in(0,1)$,
\[
    \{p_j^{\mathrm{SL}}\le z\}
    \subseteq
    \{T_N\le\Phi^{-1}(z)\}.
\]
Using Eq.~(\ref{eq:app_studentized_clt}),
\[
\begin{aligned}
    \limsup_{N\to\infty}
    \Prob_{H_j}\!\left(
        p_j^{\mathrm{SL}}\le z
    \right)
    &\le
    \lim_{N\to\infty}
    \Prob\!\left(
        T_N\le\Phi^{-1}(z)
    \right) =
    \Phi\!\left(
        \Phi^{-1}(z)
    \right)
    =z.
\end{aligned}
\]
The endpoints $z=0$ and $z=1$ are immediate. Hence
$p_j^{\mathrm{SL}}$ is asymptotically super-uniform under
$H_j:R_{\mathrm{pred},N}(\theta_j)>\alpha$.
\end{proof}

\subsection{Proof of Predictable-Variance Consistency}
\label{app:variance_estimation}

\begin{proof}[Proof of Lemma~\ref{lem:variance_consistency}]
Fix $\theta_j\in\Theta$ and suppress the argument $\theta_j$. Since
$L_n\in\{0,1\}$,
\[
    \E[
        Z_n^2
        \mid
        \mathcal F_{n-1}
    ]
    =
    p_n(1-p_n).
\]
Therefore,
\[
    V_N^2
    =
    \sum_{n=1}^{N}
    p_n(1-p_n),
\]
while Eq.~(\ref{eq:vhat_plugin}) gives
\[
    \widehat V_N^2
    =
    \sum_{n=1}^{N}
    \widehat p_n(1-\widehat p_n).
\]

Let $g(x):=x(1-x)$ for $x\in[0,1]$. Since
$|g(x)-g(y)|\le|x-y|$ for all $x,y\in[0,1]$,
\[
\begin{aligned}
    \left|
        \frac{
            \widehat V_N^2-V_N^2
        }{
            N
        }
    \right|
    &\le
    \frac1N
    \sum_{n=1}^{N}
    |\widehat p_n-p_n|\le
    \left[
        \frac1N
        \sum_{n=1}^{N}
        (\widehat p_n-p_n)^2
    \right]^{1/2}
    =
    o_p(1),
\end{aligned}
\]
where the second inequality follows from Cauchy--Schwarz.

By the assumed nondegenerate variance limit
$V_N^2/N\xrightarrow{p}\sigma_j^2>0$,
\[
    \frac{
        \widehat V_N^2
    }{
        V_N^2
    }
    -1
    =
    \frac{
        (\widehat V_N^2-V_N^2)/N
    }{
        V_N^2/N
    }
    \xrightarrow{p}0.
\]
Hence
$\widehat V_N^2/V_N^2\xrightarrow{p}1$, and the continuous mapping theorem
gives
\[
    \frac{
        \widehat V_N
    }{
        V_N
    }
    \xrightarrow{p}1.
\]

Finally, define $v_{N,j}^2:=N\sigma_j^2$. Then
$v_{N,j}^2\to\infty$ and
\[
    \frac{
        V_N^2
    }{
        v_{N,j}^2
    }
    =
    \frac{
        V_N^2/N
    }{
        \sigma_j^2
    }
    \xrightarrow{p}1.
\]
Thus Assumption~\ref{assum:lindeberg} also holds.
\end{proof}

\subsection{Proof of Simultaneous StreamLTT Certification}
\label{app:proof_certified_region}

\begin{proof}[Proof of Theorem~\ref{thm:certified_region}]
At horizon $N$, define the unsafe-policy index set
\[
    \mathcal H_{0,N}
    :=
    \left\{
        j\in\{1,\ldots,M\}:
        R_{\mathrm{pred},N}(\theta_j)>\alpha
    \right\},
\]
and let $\mathcal R_N$ denote the rejection set returned by the
multiple-testing procedure. By definition,
\[
    \widehat\Theta_{\safe}
    =
    \{\theta_j:j\in\mathcal R_N\},
    \qquad
    \Theta_{\safe,N}
    =
    \left\{
        \theta_j:
        R_{\mathrm{pred},N}(\theta_j)\le\alpha
    \right\}.
\]

Under this convention, false certification occurs exactly when at least one
unsafe policy is rejected:
\begin{equation}
    \left\{
        \widehat\Theta_{\safe}
        \nsubseteq
        \Theta_{\safe,N}
    \right\}
    =
    \left\{
        \mathcal R_N
        \cap
        \mathcal H_{0,N}
        \neq\emptyset
    \right\}.
    \label{eq:app_certification_fwer}
\end{equation}

Suppose first that the policy-level $p$-values are finite-sample
super-uniform under their corresponding null hypotheses and that the
multiple-testing procedure controls the FWER at level $\delta$. Then
\[
    \Prob\!\left(
        \mathcal R_N
        \cap
        \mathcal H_{0,N}
        \neq\emptyset
    \right)
    \le
    \delta.
\]
Together with Eq.~(\ref{eq:app_certification_fwer}), this gives
\[
    \Prob\!\left(
        \widehat\Theta_{\safe}
        \nsubseteq
        \Theta_{\safe,N}
    \right)
    \le
    \delta,
\]
or equivalently,
\[
    \Prob\!\left(
        \widehat\Theta_{\safe}
        \subseteq
        \Theta_{\safe,N}
    \right)
    \ge
    1-\delta.
\]

We next establish the post-certification statement. Let $\widehat\theta$ be
any possibly data-dependent selection satisfying
$\widehat\theta\in\widehat\Theta_{\safe}$ whenever
$\widehat\Theta_{\safe}\neq\emptyset$. On the event
$\widehat\Theta_{\safe}\subseteq\Theta_{\safe,N}$, every policy in the
certified region satisfies the predictable-risk constraint. Therefore,
\begin{equation}
    \left\{
        \widehat\Theta_{\safe}\neq\emptyset,\,
        R_{\mathrm{pred},N}(\widehat\theta)>\alpha
    \right\}
    \subseteq
    \left\{
        \widehat\Theta_{\safe}
        \nsubseteq
        \Theta_{\safe,N}
    \right\}.
    \label{eq:app_postselection_inclusion}
\end{equation}
It follows that
\[
    \Prob\!\left(
        \widehat\Theta_{\safe}\neq\emptyset,\,
        R_{\mathrm{pred},N}(\widehat\theta)>\alpha
    \right)
    \le
    \delta.
\]
This proves both finite-sample statements.

For the variance-adaptive StreamLTT test, consider Bonferroni correction,
which rejects candidate $j$ whenever
$p_j^{\mathrm{SL}}\le\delta/M$. By the union bound,
\begin{align}
    \Prob\!\left(
        \mathcal R_N
        \cap
        \mathcal H_{0,N}
        \neq\emptyset
    \right)
    &\le
    \sum_{j=1}^{M}
    \Prob\!\left(
        j\in\mathcal H_{0,N},\,
        p_j^{\mathrm{SL}}\le\frac{\delta}{M}
    \right).
    \label{eq:app_asymptotic_union}
\end{align}

For each fixed candidate $\theta_j$, Theorem~\ref{thm:lf} gives
\[
    \limsup_{N\to\infty}
    \Prob\!\left(
        j\in\mathcal H_{0,N},\,
        p_j^{\mathrm{SL}}\le\frac{\delta}{M}
    \right)
    \le
    \frac{\delta}{M}.
\]
Since $M$ is fixed and finite, taking the limit superior in
Eq.~(\ref{eq:app_asymptotic_union}) yields
\[
\begin{aligned}
    \limsup_{N\to\infty}
    \Prob\!\left(
        \mathcal R_N
        \cap
        \mathcal H_{0,N}
        \neq\emptyset
    \right)
    &\le
    \sum_{j=1}^{M}
    \limsup_{N\to\infty}
    \Prob\!\left(
        j\in\mathcal H_{0,N},\,
        p_j^{\mathrm{SL}}\le\frac{\delta}{M}
    \right) \le
    \sum_{j=1}^{M}
    \frac{\delta}{M}
    =
    \delta.
\end{aligned}
\]

Using Eq.~(\ref{eq:app_certification_fwer}),
\[
    \limsup_{N\to\infty}
    \Prob\!\left(
        \widehat\Theta_{\safe}
        \nsubseteq
        \Theta_{\safe,N}
    \right)
    \le
    \delta,
\]
which is equivalent to
\[
    \liminf_{N\to\infty}
    \Prob\!\left(
        \widehat\Theta_{\safe}
        \subseteq
        \Theta_{\safe,N}
    \right)
    \ge
    1-\delta.
\]

Finally, Eq.~(\ref{eq:app_postselection_inclusion}) holds for every $N$.
Hence
\[
\begin{aligned}
    &\limsup_{N\to\infty}
    \Prob\!\left(
        \widehat\Theta_{\safe}\neq\emptyset,\,
        R_{\mathrm{pred},N}(\widehat\theta)>\alpha
    \right) \le
    \limsup_{N\to\infty}
    \Prob\!\left(
        \widehat\Theta_{\safe}
        \nsubseteq
        \Theta_{\safe,N}
    \right)
    \le
    \delta.
\end{aligned}
\]
This proves both asymptotic statements.
\end{proof}

%% file: Expeiments_detail.tex
\section{Experimental Details}
\label{app:experimental_details}

This section provides additional details for the experiments in
Section~\ref{sec:experiments}, including the forecasting protocol,
construction of the deployment risks, certification methods, and evaluation
metrics. All model configurations, auxiliary quantities, and test specific
choices are fixed before certification.

\subsection{Datasets and Forecasting Protocol}
\label{app:forecasting_protocol}

We consider seven real world forecasting datasets spanning energy, traffic,
and temperature forecasting. The energy datasets are
Electricity~\citep{trindade2015electricity},
Appliances Energy~\citep{candanedo2017data}, and
Household Power~\citep{hebrail2006household}; the traffic datasets are
PEMS04 and PEMS08~\citep{guo2019attention}; and the temperature datasets are
Beijing Weather~\citep{chen2017beijing} and
Jena Weather~\citep{jena_climate}. The datasets differ in sequence length,
sampling frequency, and temporal structure, while their chronological order
is preserved throughout preprocessing, forecasting, and certification.

Each dataset is split chronologically into a training set (20\%), a
validation set (2\%), a calibration set (8\%), and a certification stream
(70\%). A temporal convolutional network (TCN) is trained for one step
forecasting on the training set, selected using the validation set, and then
frozen. The calibration set determines auxiliary quantities required by the
deployment risks, such as severe miss tolerances and regime thresholds,
which remain fixed throughout certification. The final 70\% of observations
form the certification stream and are processed sequentially in their
original temporal order.

\textbf{Forecasting model.}
We use the same TCN forecasting architecture across all seven datasets. The
model takes the previous 24 observations of the target variable as input and
predicts the next outcome; Electricity additionally includes temporal and
customer information. Inputs and targets are standardized using statistics
from the training set. Models are trained by minimizing mean squared error
with validation based early stopping, and the selected model is frozen
before calibration and certification. The resulting forecasts are shared
across all certification methods.

At each certification round, predictions and deployment decisions use only
information available before the corresponding outcome is revealed. History
dependent decisions may additionally use previously revealed outcomes,
whereas unrevealed outcomes remain unavailable.

Table~\ref{tab:dataset_summary} summarizes the application domains and
forecasting tasks.

\begin{table}[t]
\centering
\caption{Summary of the forecasting datasets used in the experiments.}
\label{tab:dataset_summary}
\small
\setlength{\tabcolsep}{7pt}
\renewcommand{\arraystretch}{1.05}
\begin{tabular}{lll}
\toprule
Dataset & Domain & Forecasting task \\
\midrule
Electricity     & Energy      & Load forecasting \\
Appliances      & Energy      & Energy use forecasting \\
Household Power & Energy      & Power forecasting \\
PEMS04          & Traffic     & Traffic forecasting \\
PEMS08          & Traffic     & Traffic forecasting \\
Beijing         & Temperature & Weather forecasting \\
Jena            & Temperature & Weather forecasting \\
\bottomrule
\end{tabular}
\end{table}
\subsection{Deployment Risk Construction}
\label{app:risk_construction}

We consider the seven binary deployment risks summarized in
Table~\ref{tab:risk_definitions}. They span pointwise forecasting failures,
regime dependent events, directional decisions, and history dependent
reversal errors. Each loss takes values in $\{0,1\}$, matching the Bernoulli
setting used for predictable variance estimation in
Section~\ref{sec:statistical_certification}.

For the interval based risks, let $\widehat\mu_i$ denote the frozen point
forecast and define
\[
    C_m(X_i)
    =
    [\widehat\mu_i-ms_i,\,
     \widehat\mu_i+ms_i],
\]
where $s_i$ is a calibration based residual scale fixed before certification,
and $m$ indexes the candidate interval widths.

\textbf{Interval miss.}
The interval miss loss is
$L_i^{\mathrm{int}}(m)=\ind\{Y_i\notin C_m(X_i)\}$,
with target $\alpha=0.10$.

\textbf{Lower tail miss.}
The lower tail loss isolates outcomes below the lower prediction limit,
$L_i^{\mathrm{low}}(m)=\ind\{Y_i<\ell_i(m)\}$,
with $\alpha=0.10$.

\textbf{Severe interval miss.}
Let $e_0$ denote an additional tolerance determined from the calibration
set. We define
$L_i^{\mathrm{sev}}(m)
=\ind\{|Y_i-\widehat\mu_i|>ms_i+e_0\}$,
with $\alpha=0.10$. This risk captures errors exceeding the ordinary
uncertainty scale by the prescribed tolerance.

\textbf{Task regime miss.}
Let $G_i\in\{0,1\}$ indicate an extreme operating regime determined from the
calibration set. The corresponding loss is
$L_i^{\mathrm{reg}}(m)
=\ind\{G_i=1,\;Y_i\notin C_m(X_i)\}$,
with $\alpha=0.05$. This is a joint event risk rather than miscoverage
conditional on $G_i=1$.

\textbf{Opposite direction.}
Let $a_i(\theta)\in\{-1,0,1\}$ denote the action issued by policy $\theta$
and $z_i^d\in\{-1,0,1\}$ the realized direction. We define
$L_i^{\mathrm{dir}}(\theta)
=\ind\{a_i(\theta)z_i^d=-1\}$,
with $\alpha=0.10$.

\textbf{Missed reversal.}
Let $r_i\in\{0,1\}$ denote the realized trend reversal indicator and
$\widehat r_i(\theta)\in\{0,1\}$ the corresponding policy prediction. The
loss is
$L_i^{\mathrm{miss}}(\theta)
=\ind\{r_i=1,\widehat r_i(\theta)=0\}$,
with $\alpha=0.10$.

\textbf{False reversal.}
The complementary false alert loss is
$L_i^{\mathrm{false}}(\theta)
=\ind\{r_i=0,\widehat r_i(\theta)=1\}$,
with $\alpha=0.10$. Together, the missed and false reversal risks distinguish
the two asymmetric failure modes of reversal detection.

All quantities defining these risks are either fixed using precertification
data or constructed from information available before the corresponding
outcome is revealed.

\subsection{Certification Methods}
\label{app:certification_methods}

Our primary implementation, \emph{StreamLTT-A}, uses the variance adaptive
test in Theorem~\ref{thm:lf}. We compare it with five baselines:
\emph{IID-LTT}~\citep{angelopoulos2021ltt},
\emph{LTT-HB-Holm}~\citep{hoeffding1963probability,bentkus2004hoeffding},
\emph{HAC-CLT-Holm}~\citep{newey1987simple},
\emph{CircularBlockBootstrap-Holm}~\citep{politis1992circular}, and
\emph{EB-Howard-Bonferroni}~\citep{howard2020timeuniform}.
We additionally evaluate \emph{StreamLTT-H}, which uses the same average
predictable risk target as StreamLTT-A but replaces the adaptive test with
the finite sample test in Theorem~\ref{thm:hoeffding_inid}.

IID-LTT and LTT-HB-Holm both retain the standard LTT population risk under
independent calibration losses. IID-LTT uses a CLT approximation to
empirical risk, whereas LTT-HB-Holm uses the finite sample
Hoeffding--Bentkus bounded loss test. HAC-CLT-Holm instead adjusts the long
run variance for temporal dependence, while CircularBlockBootstrap-Holm
approximates the dependent sampling distribution through circular block
resampling. EB-Howard-Bonferroni uses an empirical Bernstein concentration
construction.

For StreamLTT-A, the one step predictable risk
$p_n(\theta)=\E[L_n(\theta)\mid\mathcal F_{n-1}]$ is estimated using only
information available before $Y_n$ is revealed. Given
$\widehat p_n(\theta)$, the predictable quadratic variation is estimated as
\[
    \widehat V_N^2(\theta)
    =
    \sum_{n=1}^{N}
    \widehat p_n(\theta)
    \bigl(1-\widehat p_n(\theta)\bigr),
\]
as in Eq.~\eqref{eq:vhat_plugin}. Pointwise risks use the current forecasting
information and policy quantities, while regime and history dependent risks
add the corresponding information available in $\mathcal F_{n-1}$. All
estimator configurations are fixed before certification.

StreamLTT-H retains the same predictable risk formulation but uses the
finite sample martingale Hoeffding $p$ value in
Eq.~\eqref{eq:hoeffding_pvalue}. Comparing StreamLTT-H with StreamLTT-A
therefore isolates predictable variance adaptation while holding the
certified risk target fixed.

\subsection{Evaluation Protocol and Metrics}
\label{app:evaluation_protocol}

We evaluate all $7\times7=49$ dataset--risk combinations at
$\delta\in\{0.10,0.20\}$ using $2{,}000$ independent repetitions per
configuration.

A family wise error occurs when at least one certified policy exceeds the
prescribed predictable risk target,
\[
    \exists\,
    \theta\in\widehat\Theta_{\safe}
    \quad\text{such that}\quad
    R_{\mathrm{pred},N}(\theta)>\alpha.
\]
Empirical FWER is the frequency of this event across repetitions.

For each nonempty certified region, define
\[
    R_{\max}
    :=
    \max_{\theta\in\widehat\Theta_{\safe}}
    R_{\mathrm{pred},N}(\theta),
\]
and let $\overline R_{\max}$ denote its mean across nonempty repetitions.
We measure certification tightness by the risk gap
$\alpha-\overline R_{\max}$ and family wise budget utilization by the FWER
gap $\delta-\widehat{\mathrm{FWER}}$. Values near zero indicate that the
corresponding certification boundary is approached closely.

As a complementary measure, the certified fraction
$|\widehat\Theta_{\safe}|/|\Theta|$ records the proportion of candidate
policies retained after certification. Monte Carlo uncertainty is computed
from the $2{,}000$ repetitions. For empirical FWER $\widehat q$, its Monte
Carlo standard error is
$\sqrt{\widehat q(1-\widehat q)/2000}$; we additionally compute binomial
confidence intervals for FWER and Monte Carlo uncertainty for
$\overline R_{\max}$ and the certified fraction.

%% file: Additional_Exp.tex
\section{Additional Experimental Results}
\label{app:additional_results}

This section complements Section~\ref{sec:results} with complete dataset
level results for all seven deployment risks at both family wise levels,
the finite sample StreamLTT ablation, and Monte Carlo diagnostics.

\subsection{Dataset Level Results at $\delta=0.10$}
\label{app:dataset_results_010}

Table~\ref{tab:dataset_results_010} reports the complete dataset level
comparison at $\delta=0.10$. Each entry gives
$\overline R_{\max}/\widehat{\mathrm{FWER}}$, allowing frontier tightness
and false certification to be compared jointly across datasets and risks.

\begin{table*}[t]
\centering
\caption{
Dataset level certification results at $\delta=0.10$ for all seven
deployment risks. Each entry reports
$\overline R_{\max}/\widehat{\mathrm{FWER}}$.
}
\label{tab:dataset_results_010}
\scriptsize
\setlength{\tabcolsep}{2.0pt}
\renewcommand{\arraystretch}{1.04}
\begin{tabular}{llccccccc}
\toprule
Risk & Method
& Electricity & Appliances & Household
& PEMS04 & PEMS08 & Beijing & Jena \\
\midrule

\multirow{6}{*}{\shortstack[l]{Interval miss\\($\alpha=.10$)}}
& IID-LTT
& .051/.028 & .046/.013 & .048/.020 & .051/.017
& .044/.015 & .047/.019 & .045/.012 \\
& LTT-HB-Holm
& .050/.006 & .045/.004 & .047/.009 & .051/.006
& .043/.006 & .046/.005 & .045/.004 \\
& HAC-CLT-Holm
& .051/.035 & .046/.014 & .048/.025 & .051/.021
& .044/.019 & .047/.024 & .045/.017 \\
& BlockBootstrap-Holm
& .051/.034 & .046/.015 & .048/.025 & .052/.021
& .044/.019 & .047/.023 & .045/.018 \\
& EB-Howard-Bonf.
& .050/.000 & .045/.000 & .047/.000 & .050/.000
& .043/.000 & .046/.000 & .044/.000 \\
& \textbf{StreamLTT-A}
& \textbf{.096/.084} & \textbf{.096/.082} & \textbf{.097/.102}
& \textbf{.096/.094} & \textbf{.095/.087} & \textbf{.096/.089}
& \textbf{.096/.083} \\
\addlinespace[2pt]

\multirow{6}{*}{\shortstack[l]{Lower tail miss\\($\alpha=.10$)}}
& IID-LTT
& .055/.045 & .069/.031 & .053/.032 & .065/.025
& .066/.026 & .069/.020 & .064/.025 \\
& LTT-HB-Holm
& .053/.014 & .069/.010 & .052/.009 & .065/.009
& .066/.009 & .069/.004 & .063/.007 \\
& HAC-CLT-Holm
& .055/.052 & .069/.037 & .053/.037 & .066/.030
& .066/.031 & .070/.023 & .064/.032 \\
& BlockBootstrap-Holm
& .055/.052 & .069/.036 & .053/.036 & .066/.031
& .066/.031 & .070/.023 & .064/.031 \\
& EB-Howard-Bonf.
& .052/.000 & .068/.000 & .052/.000 & .065/.000
& .065/.000 & .069/.000 & .063/.000 \\
& \textbf{StreamLTT-A}
& \textbf{.097/.095} & \textbf{.096/.097} & \textbf{.095/.093}
& \textbf{.097/.080} & \textbf{.097/.088} & \textbf{.096/.081}
& \textbf{.097/.085} \\
\addlinespace[2pt]

\multirow{6}{*}{\shortstack[l]{Severe miss\\($\alpha=.10$)}}
& IID-LTT
& .054/.083 & .080/.082 & .063/.079 & .075/.090
& .069/.091 & .072/.097 & .073/.096 \\
& LTT-HB-Holm
& .051/.029 & .078/.026 & .060/.024 & .073/.029
& .066/.028 & .070/.030 & .071/.031 \\
& HAC-CLT-Holm
& .054/.092 & .080/.094 & .063/.087 & .075/.097
& .069/.099 & .073/.105 & .073/.107 \\
& BlockBootstrap-Holm
& .054/.091 & .080/.093 & .063/.086 & .075/.096
& .069/.098 & .073/.105 & .073/.105 \\
& EB-Howard-Bonf.
& .050/.000 & .078/.000 & .059/.000 & .072/.000
& .065/.000 & .069/.000 & .070/.000 \\
& \textbf{StreamLTT-A}
& \textbf{.072/.085} & \textbf{.086/.082} & \textbf{.063/.080}
& \textbf{.075/.090} & \textbf{.069/.091} & \textbf{.072/.097}
& \textbf{.073/.096} \\
\addlinespace[2pt]

\multirow{6}{*}{\shortstack[l]{Task regime miss\\($\alpha=.05$)}}
& IID-LTT
& .025/.097 & .016/.016 & .032/.017 & .040/.048
& .040/.059 & .025/.015 & .038/.050 \\
& LTT-HB-Holm
& .024/.033 & .016/.005 & .032/.005 & .040/.017
& .040/.018 & .025/.004 & .038/.017 \\
& HAC-CLT-Holm
& .026/.099 & .016/.018 & .032/.021 & .040/.049
& .040/.060 & .025/.018 & .038/.051 \\
& BlockBootstrap-Holm
& .026/.100 & .016/.017 & .032/.021 & .040/.048
& .040/.060 & .025/.017 & .038/.050 \\
& EB-Howard-Bonf.
& .023/.000 & .016/.000 & .032/.000 & .039/.000
& .040/.000 & .025/.000 & .038/.000 \\
& \textbf{StreamLTT-A}
& \textbf{.034/.097} & \textbf{.044/.117} & \textbf{.047/.086}
& \textbf{.046/.093} & \textbf{.045/.105} & \textbf{.048/.084}
& \textbf{.045/.096} \\
\addlinespace[2pt]

\multirow{6}{*}{\shortstack[l]{Opposite direction\\($\alpha=.10$)}}
& IID-LTT
& .095/.085 & .094/.086 & .083/.085 & .088/.087
& .091/.088 & .074/.088 & .054/.089 \\
& LTT-HB-Holm
& .095/.032 & .094/.029 & .082/.027 & .088/.026
& .090/.029 & .072/.026 & .052/.034 \\
& HAC-CLT-Holm
& .095/.095 & .094/.097 & .083/.091 & .089/.099
& .091/.098 & .074/.101 & .055/.099 \\
& BlockBootstrap-Holm
& .095/.097 & .094/.096 & .084/.092 & .089/.098
& .091/.098 & .074/.098 & .055/.099 \\
& EB-Howard-Bonf.
& .087/.000 & .091/.000 & .082/.000 & .087/.000
& .090/.000 & .071/.000 & .050/.000 \\
& \textbf{StreamLTT-A}
& \textbf{.097/.087} & \textbf{.094/.087} & \textbf{.083/.086}
& \textbf{.097/.087} & \textbf{.097/.088} & \textbf{.092/.090}
& \textbf{.055/.091} \\
\addlinespace[2pt]

\multirow{6}{*}{\shortstack[l]{Missed reversal\\($\alpha=.10$)}}
& IID-LTT
& .080/.087 & .064/.026 & .075/.040 & .048/.016
& .050/.018 & .089/.091 & .081/.047 \\
& LTT-HB-Holm
& .077/.030 & .063/.008 & .074/.015 & .048/.005
& .049/.006 & .086/.031 & .080/.018 \\
& HAC-CLT-Holm
& .081/.099 & .064/.033 & .075/.045 & .049/.022
& .050/.023 & .090/.099 & .081/.052 \\
& BlockBootstrap-Holm
& .081/.097 & .064/.033 & .075/.047 & .049/.022
& .050/.023 & .090/.099 & .081/.053 \\
& EB-Howard-Bonf.
& .073/.000 & .063/.000 & .074/.000 & .048/.001
& .049/.000 & .079/.000 & .080/.000 \\
& \textbf{StreamLTT-A}
& \textbf{.097/.088} & \textbf{.095/.092} & \textbf{.095/.085}
& \textbf{.095/.090} & \textbf{.095/.089} & \textbf{.096/.093}
& \textbf{.082/.092} \\
\addlinespace[2pt]

\multirow{6}{*}{\shortstack[l]{False reversal\\($\alpha=.10$)}}
& IID-LTT
& .089/.088 & .094/.086 & .084/.081 & .049/.018
& .045/.020 & .091/.105 & .076/.098 \\
& LTT-HB-Holm
& .089/.029 & .094/.033 & .083/.028 & .048/.004
& .044/.006 & .090/.028 & .074/.030 \\
& HAC-CLT-Holm
& .090/.098 & .094/.096 & .084/.091 & .049/.022
& .045/.023 & .091/.110 & .076/.111 \\
& BlockBootstrap-Holm
& .090/.100 & .094/.096 & .084/.089 & .049/.022
& .045/.024 & .091/.110 & .076/.107 \\
& EB-Howard-Bonf.
& .088/.000 & .090/.000 & .082/.000 & .048/.000
& .044/.000 & .090/.000 & .073/.000 \\
& \textbf{StreamLTT-A}
& \textbf{.097/.088} & \textbf{.097/.086} & \textbf{.096/.081}
& \textbf{.096/.086} & \textbf{.096/.088} & \textbf{.098/.105}
& \textbf{.094/.100} \\
\bottomrule
\end{tabular}
\end{table*}

The complete results show that the aggregate behavior in
Section~\ref{sec:results} is not driven by a single dataset or risk. The
largest separation appears for interval and lower tail miss, where
StreamLTT-A consistently moves the frontier close to $\alpha=0.10$ while
the competing methods remain substantially below it. Similar gains appear
for missed and false reversal, particularly on datasets where the baseline
frontier is far from the target.

Other risks illustrate why frontier tightness and FWER must be interpreted
jointly. For severe miss and opposite direction, HAC-CLT-Holm and the block
bootstrap often attain empirical FWER near $0.10$, but their frontiers remain
close to IID-LTT on several datasets. Conversely, StreamLTT-A maintains FWER
near the prescribed level while improving the frontier where additional
admissible policies are available. The attainable frontier can itself vary
by dataset, as seen for opposite direction on Jena and for severe miss on
several streams.

\subsection{Dataset Level Results at $\delta=0.20$}
\label{app:dataset_results_020}

Table~\ref{tab:dataset_results_020} reports the corresponding dataset level
results after increasing the family wise level to $\delta=0.20$. This
comparison shows how each procedure responds when a larger false
certification budget is available.

\begin{table*}[t]
\centering
\caption{
Dataset level certification results at $\delta=0.20$.
Each entry reports
$\overline R_{\max}/\widehat{\mathrm{FWER}}$.
}
\label{tab:dataset_results_020}
\scriptsize
\setlength{\tabcolsep}{2.4pt}
\renewcommand{\arraystretch}{1.04}
\begin{tabular}{llccccccc}
\toprule
Risk & Method
& Electricity & Appliances & Household
& PEMS04 & PEMS08 & Beijing & Jena \\
\midrule

\multirow{6}{*}{\shortstack[l]{Interval miss\\($\alpha=.10$)}}
& IID-LTT
& .052/.054 & .046/.022 & .049/.038 & .052/.035
& .044/.032 & .048/.033 & .046/.029 \\
& LTT-HB-Holm
& .050/.016 & .046/.009 & .048/.016 & .051/.011
& .043/.011 & .046/.011 & .045/.008 \\
& HAC-CLT-Holm
& .053/.060 & .047/.029 & .049/.046 & .053/.044
& .045/.041 & .048/.038 & .046/.038 \\
& BlockBootstrap-Holm
& .053/.061 & .047/.028 & .049/.046 & .053/.044
& .045/.042 & .048/.038 & .046/.038 \\
& EB-Howard-Bonf.
& .050/.000 & .045/.000 & .047/.000 & .050/.000
& .043/.000 & .046/.000 & .044/.000 \\
& \textbf{StreamLTT-A}
& \textbf{.097/.185} & \textbf{.096/.178} & \textbf{.097/.211}
& \textbf{.097/.186} & \textbf{.095/.181} & \textbf{.097/.177}
& \textbf{.097/.182} \\
\midrule

\multirow{6}{*}{\shortstack[l]{Lower tail miss\\($\alpha=.10$)}}
& IID-LTT
& .057/.095 & .070/.063 & .055/.062 & .066/.050
& .067/.051 & .070/.035 & .065/.059 \\
& LTT-HB-Holm
& .054/.030 & .069/.021 & .052/.017 & .065/.018
& .066/.018 & .069/.014 & .063/.018 \\
& HAC-CLT-Holm
& .057/.102 & .070/.070 & .055/.068 & .067/.058
& .068/.065 & .070/.044 & .065/.064 \\
& BlockBootstrap-Holm
& .057/.101 & .070/.070 & .055/.068 & .067/.057
& .068/.065 & .070/.043 & .065/.065 \\
& EB-Howard-Bonf.
& .052/.000 & .068/.000 & .052/.000 & .065/.000
& .065/.000 & .069/.000 & .063/.000 \\
& \textbf{StreamLTT-A}
& \textbf{.098/.184} & \textbf{.096/.183} & \textbf{.096/.199}
& \textbf{.098/.175} & \textbf{.098/.185} & \textbf{.097/.175}
& \textbf{.097/.184} \\
\midrule

\multirow{6}{*}{\shortstack[l]{Severe miss\\($\alpha=.10$)}}
& IID-LTT
& .059/.180 & .082/.173 & .067/.179 & .078/.188
& .072/.189 & .075/.195 & .076/.185 \\
& LTT-HB-Holm
& .053/.056 & .079/.057 & .062/.058 & .074/.067
& .068/.066 & .071/.067 & .072/.063 \\
& HAC-CLT-Holm
& .060/.194 & .082/.186 & .067/.191 & .078/.202
& .072/.198 & .076/.202 & .076/.199 \\
& BlockBootstrap-Holm
& .059/.192 & .082/.185 & .067/.188 & .078/.199
& .072/.197 & .076/.203 & .076/.198 \\
& EB-Howard-Bonf.
& .050/.000 & .078/.000 & .059/.000 & .072/.000
& .065/.000 & .069/.000 & .070/.000 \\
& \textbf{StreamLTT-A}
& \textbf{.075/.185} & \textbf{.088/.175} & \textbf{.067/.180}
& \textbf{.078/.188} & \textbf{.072/.190} & \textbf{.075/.196}
& \textbf{.076/.189} \\
\midrule

\multirow{6}{*}{\shortstack[l]{Task regime\\($\alpha=.05$)}}
& IID-LTT
& .028/.189 & .017/.033 & .032/.034 & .040/.093
& .041/.105 & .026/.032 & .039/.096 \\
& LTT-HB-Holm
& .025/.064 & .016/.007 & .032/.010 & .040/.029
& .040/.038 & .025/.009 & .038/.034 \\
& HAC-CLT-Holm
& .028/.190 & .017/.035 & .033/.038 & .040/.098
& .041/.111 & .026/.034 & .039/.099 \\
& BlockBootstrap-Holm
& .028/.188 & .017/.035 & .033/.040 & .040/.096
& .041/.108 & .026/.034 & .039/.100 \\
& EB-Howard-Bonf.
& .023/.000 & .016/.000 & .032/.000 & .039/.000
& .040/.000 & .025/.000 & .038/.000 \\
& \textbf{StreamLTT-A}
& \textbf{.036/.189} & \textbf{.045/.217} & \textbf{.048/.193}
& \textbf{.046/.191} & \textbf{.045/.212} & \textbf{.048/.192}
& \textbf{.046/.197} \\
\midrule

\multirow{6}{*}{\shortstack[l]{Opposite direction\\($\alpha=.10$)}}
& IID-LTT
& .096/.187 & .095/.178 & .085/.172 & .090/.180
& .092/.190 & .077/.189 & .060/.196 \\
& LTT-HB-Holm
& .095/.061 & .094/.064 & .083/.060 & .088/.057
& .090/.063 & .073/.056 & .053/.063 \\
& HAC-CLT-Holm
& .096/.198 & .095/.191 & .085/.186 & .090/.191
& .092/.203 & .077/.199 & .060/.207 \\
& BlockBootstrap-Holm
& .096/.196 & .095/.189 & .085/.183 & .090/.190
& .092/.203 & .077/.199 & .060/.206 \\
& EB-Howard-Bonf.
& .087/.000 & .092/.000 & .082/.000 & .087/.000
& .090/.000 & .071/.000 & .050/.000 \\
& \textbf{StreamLTT-A}
& \textbf{.097/.188} & \textbf{.095/.180} & \textbf{.085/.179}
& \textbf{.098/.182} & \textbf{.098/.192} & \textbf{.093/.192}
& \textbf{.060/.200} \\
\midrule

\multirow{6}{*}{\shortstack[l]{Missed reversal\\($\alpha=.10$)}}
& IID-LTT
& .084/.173 & .065/.058 & .076/.085 & .050/.042
& .051/.041 & .092/.191 & .082/.090 \\
& LTT-HB-Holm
& .079/.061 & .064/.018 & .074/.030 & .048/.011
& .049/.011 & .088/.064 & .081/.035 \\
& HAC-CLT-Holm
& .085/.182 & .065/.066 & .076/.096 & .050/.051
& .051/.047 & .093/.206 & .082/.100 \\
& BlockBootstrap-Holm
& .084/.180 & .065/.066 & .076/.095 & .050/.051
& .051/.046 & .093/.204 & .082/.100 \\
& EB-Howard-Bonf.
& .073/.000 & .063/.000 & .074/.000 & .048/.001
& .049/.000 & .079/.000 & .080/.000 \\
& \textbf{StreamLTT-A}
& \textbf{.098/.173} & \textbf{.095/.191} & \textbf{.095/.192}
& \textbf{.096/.186} & \textbf{.096/.185} & \textbf{.097/.195}
& \textbf{.084/.201} \\
\midrule

\multirow{6}{*}{\shortstack[l]{False reversal\\($\alpha=.10$)}}
& IID-LTT
& .091/.183 & .095/.178 & .085/.163 & .050/.043
& .046/.039 & .092/.209 & .079/.199 \\
& LTT-HB-Holm
& .089/.066 & .094/.058 & .083/.057 & .048/.010
& .045/.012 & .091/.075 & .075/.070 \\
& HAC-CLT-Holm
& .091/.197 & .095/.193 & .085/.178 & .050/.050
& .047/.043 & .092/.226 & .079/.210 \\
& BlockBootstrap-Holm
& .091/.195 & .095/.192 & .085/.177 & .050/.049
& .047/.043 & .092/.224 & .079/.208 \\
& EB-Howard-Bonf.
& .088/.000 & .091/.000 & .082/.000 & .048/.000
& .044/.000 & .090/.000 & .073/.000 \\
& \textbf{StreamLTT-A}
& \textbf{.098/.186} & \textbf{.097/.179} & \textbf{.097/.163}
& \textbf{.097/.192} & \textbf{.097/.198} & \textbf{.098/.214}
& \textbf{.095/.202} \\
\bottomrule
\end{tabular}
\end{table*}

The larger family wise budget produces two distinct behaviors. StreamLTT-A
responds across nearly all dataset--risk pairs: empirical FWER generally
moves toward $0.20$, while the certified frontier remains close to the
corresponding risk target whenever the candidate family contains policies
near that boundary. For example, the interval and lower tail frontiers
remain around $0.095$--$0.098$ across all seven datasets, while their
empirical FWER values move from the neighborhood of $0.10$ at
$\delta=0.10$ toward $0.20$.

The baseline response is less uniform. HAC-CLT-Holm and
BlockBootstrap-Holm approach the larger FWER budget for some risks, such as
severe miss and opposite direction, but remain substantially below it for
interval, lower tail, task regime, and several reversal settings.
LTT-HB-Holm remains more conservative throughout, while
EB-Howard-Bonferroni leaves essentially the entire family wise budget unused.
Moreover, increasing $\delta$ changes many baseline frontiers only modestly.
For interval miss, for example, their $\overline R_{\max}$ values remain
around $0.04$--$0.05$, whereas StreamLTT-A remains near $0.10$.

\subsection{Finite Sample StreamLTT Ablation}
\label{app:hoeffding_ablation}

To complement the aggregate ablation in
Section~\ref{sec:results}, Table~\ref{tab:streamltt_h_ablation} reports the
risk wise results for StreamLTT-H after averaging across the seven datasets.
StreamLTT-H retains the same average predictable risk target as StreamLTT-A
but uses the finite sample martingale Hoeffding test, isolating the effect of
removing predictable variance adaptation.

\begin{table}[t]
\centering
\caption{
Risk wise results for StreamLTT-H, averaged across the seven datasets.
$\overline R_{\max}$ denotes the mean maximum predictable risk, FWER the
empirical family wise error rate, and Frac. the mean certified fraction.
}
\label{tab:streamltt_h_ablation}
\footnotesize
\setlength{\tabcolsep}{5pt}
\renewcommand{\arraystretch}{1.05}
\begin{tabular}{lccc|ccc}
\toprule
&
\multicolumn{3}{c}{$\delta=0.10$}
&
\multicolumn{3}{c}{$\delta=0.20$}
\\
\cmidrule(lr){2-4}
\cmidrule(lr){5-7}
Risk
& $\overline R_{\max}$ & FWER & Frac.
& $\overline R_{\max}$ & FWER & Frac. \\
\midrule
Interval miss
& .092 & .000 & .322
& .093 & .001 & .325 \\

Lower tail miss
& .094 & .000 & .699
& .094 & .001 & .704 \\

Severe miss
& .070 & .000 & .981
& .070 & .001 & .982 \\

Task regime miss
& .042 & .000 & .647
& .043 & .000 & .652 \\

Opposite direction
& .086 & .000 & .947
& .086 & .001 & .953 \\

Missed reversal
& .091 & .000 & .864
& .091 & .001 & .866 \\

False reversal
& .093 & .000 & .809
& .094 & .001 & .814 \\
\bottomrule
\end{tabular}
\end{table}

Across risks, StreamLTT-H already produces relatively tight certified
frontiers, while its empirical FWER remains essentially zero at both family
wise levels. Increasing $\delta$ from $0.10$ to $0.20$ changes the frontier
and certified fraction only modestly, consistent with the conservatism of
the finite sample Hoeffding test. Together with the StreamLTT-A results
reported above, this risk wise breakdown supports the aggregate ablation:
the predictable risk formulation accounts for most of the frontier
tightening, whereas predictable variance adaptation primarily enables the
procedure to use more of the available family wise budget.